\documentclass[runningheads]{llncs}
\usepackage{siunitx}
 
\usepackage{eccv}

\usepackage{eccvabbrv}

\usepackage{graphicx}
\usepackage{booktabs}

\usepackage[accsupp]{axessibility}  

\usepackage{hyperref}

\usepackage{orcidlink}
\definecolor{cbaseline}{HTML}{A6A6A6} 
\definecolor{cdino}{HTML}{607C8E}     
\definecolor{cdinosn}{HTML}{1A3B6E}   
\definecolor{cours}{HTML}{B02E2E}     
\definecolor{corange}{HTML}{D95F02}
\definecolor{cpurple}{HTML}{7570B3}
\usepackage{wrapfig}

\usepackage{amsmath}
\usepackage{amssymb}
\usepackage{mathtools}
\usepackage{multirow}
\usepackage{adjustbox}
\usepackage{threeparttable}
\usepackage{epsfig}
\usepackage{wrapfig}

\usepackage{algorithm}
\usepackage[table]{xcolor}
\usepackage{algorithmic}
\usepackage{setspace}
\usepackage{bm}
\usepackage{float}
\usepackage{makecell}
\usepackage{amssymb}

\usepackage{arydshln}

\usepackage{adjustbox}
\usepackage{xcolor}
\usepackage{siunitx}

\begin{document}

\title{Learning on the Manifold: Unlocking Standard Diffusion Transformers with Representation Encoders} 

\titlerunning{Abbreviated paper title}

\author{Amandeep Kumar, 
Vishal M. Patel }

\authorrunning{F.~Author et al.}

\institute{Johns Hopkins University \\
\email{\{akumar99, vpatel36\}}@jhu.edu\\
\url{https://github.com/amandpkr/RJF}
}

\maketitle

\begin{abstract}
  Leveraging representation encoders for generative modeling offers a path for efficient, high-fidelity synthesis. However, standard diffusion transformers fail to converge on these representations directly. While recent work attributes this to a capacity bottleneck—proposing computationally expensive ``width scaling'' of diffusion transformers—we demonstrate that the failure is fundamentally geometric. We identify \textit{Geometric Interference} as the root cause: standard Euclidean flow matching forces probability paths through the low-density interior of the hyperspherical feature space of representation encoders, rather than following the manifold surface. To resolve this, we propose \textbf{Riemannian Flow Matching with Jacobi Regularization (RJF)}. By constraining the generative process to the manifold geodesics and correcting for curvature-induced error propagation, RJF enables standard Diffusion Transformer architectures to converge without width scaling. Our method RJF enables the standard DiT-B architecture (131M parameters) to converge effectively, achieving an FID of \textbf{3.37} where prior methods fail to converge. 
\end{abstract}

\section{Introduction}
\label{introduction}


Flow Matching \cite{lipman2022flow, esser2024scaling, liu2022flow, albergo2022building} and Diffusion models \cite{ma2024sit, rombach2022high, ho2020denoising, song2020score} have revolutionized generative modeling, enabling high-fidelity synthesis across modalities. While initial approaches operated in pixel space, the paradigm has shifted toward Latent Diffusion Models (LDMs) \cite{rombach2022high, vahdat2021score} that leverage compressed representations of VAE \cite{kingma2013auto}. However, because VAEs are optimized for reconstruction, they predominantly capture low-level texture; this forces the diffusion model to learn high-level semantics from scratch, leading to slow convergence. To overcome this, recent works enhance the VAE latent space with semantic priors from strong representation encoders like DiNO \cite{oquab2023dinov2, simeoni2025dinov3} and SigLIP \cite{radford2021learning}. These approaches typically require explicitly aligning semantic representations within the VAE latent space \cite{yao2025reconstruction} or the diffusion intermediate features \cite{leng2025repa, yu2024representation}. Consequently, these methods requires complex auxiliary losses and additional training stages.

\begin{wrapfigure}{r}{0.6\textwidth}
    \centering
    \vspace{-17pt} 
    \includegraphics[width=\linewidth]{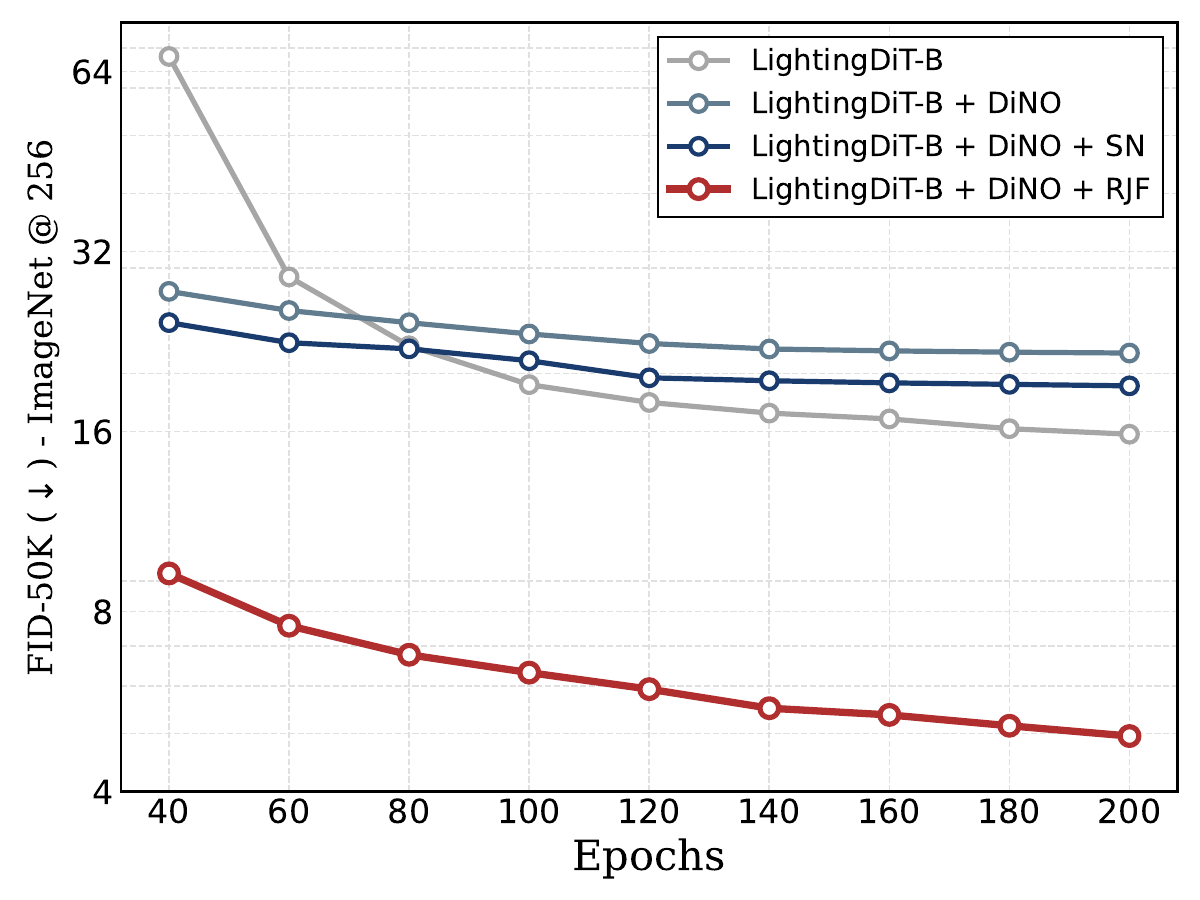}
    \vspace{-20pt} 
    \caption{\textbf{Bridging the Geometric Gap.} We demonstrate that respecting the intrinsic geometry of pre-trained representations encoders enables the use of standard Diffusion Transformers without any architectural modification such as Width Scaling \cite{zheng2025diffusion}. Our method, Riemannian Flow Matching with Jacobi Regularization (\textcolor{cours}{\textbf{+DiNO+RJF}}), achieves an FID of \textbf{4.95} using standard LightingDiT-B \cite{yao2025reconstruction} architecture without guidance, significantly outperforming the VAE-based \textcolor{cbaseline}{LightingDiT-B} (FID \textbf{15.83}). In contrast, applying standard Flow Matching to DINOv2-B features (\textcolor{cdino}{\textbf{+DiNO}}) fails to converge (FID \textbf{21.64}) due to \textit{Geometric Interference}. Even restricting the noise to the hypersphere to strictly learn the angular component (\textcolor{cdinosn}{\textbf{+DiNO+SN}}) yields only marginal improvement (FID \textbf{19.07}), as Euclidean linear paths still traverse the low-probability interior of the feature manifold.}
    \label{fig:intro_grah}
    \vspace{-23pt} 
\end{wrapfigure}

Recent work challenges this complexity by proposing Representation Autoencoders (RAE) \cite{zheng2025diffusion}, which discard the VAE entirely in favor of diffusing directly within the feature space of frozen representation encoders. RAEs demonstrate that these high-dimensional semantic representations can support high-fidelity generation without the need for auxiliary alignment losses or complex training stages. However we also have the similar observation like RAE \cite{zheng2025diffusion} 
that the standard diffusion recipe fails to converge effectively on these high-dimensional latents, even in a simplified single-image overfitting regime. While RAE attributes this failure to a capacity bottleneck, proposing to scale the transformer width to match the latent dimensionality—we identify a more fundamental cause rooted in the intrinsic geometry of the latent space. We argue that the optimization difficulty arises not from insufficient parameter count, but from a ``Geometry Gap'': a structural conflict where the Euclidean probability paths assumed by standard flow matching \cite{lipman2022flow} violate the  hyperspherical manifold of representation space.

\begin{wrapfigure}{r}{0.6\textwidth}
    \centering
    \vspace{-10pt}
    \includegraphics[width=1.0\linewidth]{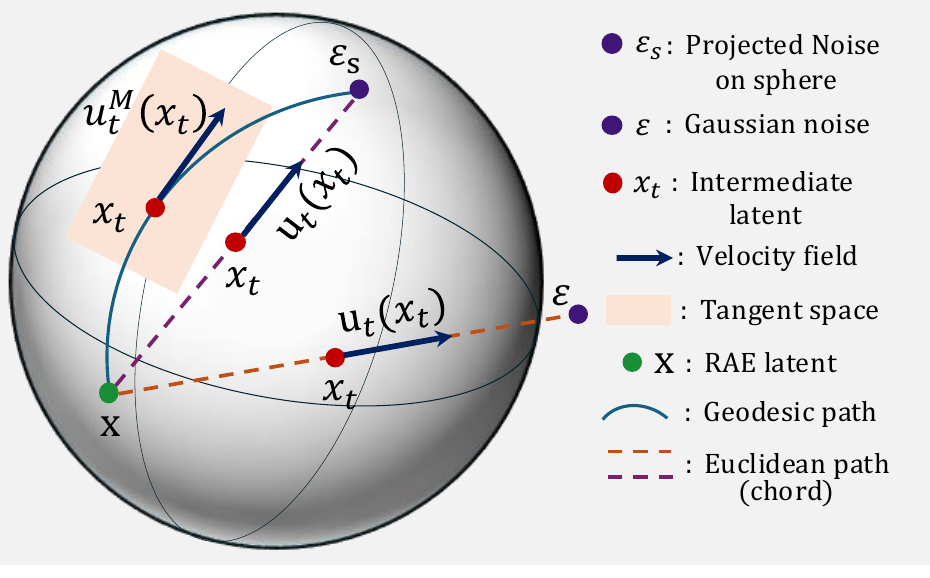}
    \vspace{-20pt}
    
   \caption{\textbf{Geometric Trajectories on the Hypersphere.} Visualization of flow matching paths on the high-dimension semantic representation manifold $\mathcal{S}^{d-1}$.
Standard Euclidean Flow Matching constructs linear paths that ignore the manifold geometry. Whether targeting standard Gaussian noise $\epsilon$ (\textcolor{corange}{\textbf{orange}}) or projecting noise onto the sphere $\epsilon_s$ (\textcolor{cpurple}{\textbf{purple}}) to strictly learn the angular component, the linear interpolation forms a chord that cuts through the low-density interior. This forces the model to learn a velocity field in undefined regions regardless of the endpoint.
In contrast,  Riemannian Flow Matching follows the geodesic \textcolor{cdinosn}{\textbf{(blue curve)}}, ensuring the intermediate state $x_t$ remains strictly on the manifold surface. The resulting velocity field $u_t^M(x_t)$ is correctly defined within the tangent space (\textcolor{pink}{\textbf{pink plane}}), naturally respecting the geometry of the high-dimsnional semantic representations.
\vspace{-10pt}
}
    \label{fig:intro_arch_2}
\end{wrapfigure}

To understand this failure, we analyze the intrinsic geometry of the feature space. We observe that these high-dimensional representations do not populate the ambient Euclidean space but are strictly confined to a hypersphere, creating a hard shell geometry, and all the information are encoded in angular vectors. We identify the root cause of convergence failure for standard diffusion transfomer as \textbf{Geometric Interference}: the standard linear probability path used in flow matching cuts through the low-density interior (off-manifold) of hypersphere (forming a chord), rather than following the manifold's surface \cite{rozen2021moser, mathieu2020riemannian}. This forces the model to learn a velocity field in regions where the representation space is undefined as shown in \cref{fig:intro_arch_2}. Crucially, we challenge the prevailing hypothesis that this requires scaling model width \cite{zheng2025diffusion}. Our experiments reveal that the model has sufficient capacity to learn the semantics, but under the standard objective, it wastes its capacity, minimizing radial error (learning feature magnitude which is fixed as the radius of hypersphere) and learning trajectories through the off-manifold area induced by this geometric mismatch.

Motivated from this insight, we propose Riemannian Flow Matching with Jacobi Regularization (RJF). First, we address the trajectory mismatch by adopting Riemannian Flow Matching \cite{chen2023flow}, which replaces the Euclidean linear path with Spherical Linear Interpolation (SLERP). This ensures the generative process follows the geodesic(shortest path along the curve between two points), staying strictly on the manifold surface. Second, we recognize that simply fixing the path is insufficient because the flow matching objective remains geometrically unaware; it treats errors uniformly. On a positively curved hypersphere, velocity errors propagate non-linearly due to the focusing of geodesics (similar to how parallel longitude lines eventually meet at the poles). To correct this, we introduce a Jacobi Regularization derived from Jacobi fields \cite{zaghen2025riemannian}, which reweights the loss to account for curvature-induced distortion. This geometric alignment enables standard DiT architectures to converge efficiently without any architectural modification.

\noindent \textbf{Our contributions are summarized as follows:}
\begin{itemize}
    \item \textbf{Geometric Analysis of Convergence Failure:} We identify Geometric Interference as fundamental bottleneck preventing standard diffusion transformers from learning directly on high-dimension representations. We demonstrate that  failure arises not only from a capacity deficit, but from the Euclidean objective forcing the model to minimize radial errors and learning trajectories through the low-density interior of the feature manifold.
    
    \item \textbf{Riemannian Flow Matching with Jacobi Regularization:} We propose a geometrical  framework that defines the generative process directly on the hyperspherical manifold of the representation encoder. By combining Riemannian Flow Matching (to correct the trajectory) with Jacobi Regularization (to account for geodesic focusing), we ensure the optimization is consistent with both the topology and curvature of the latent space.
    
    \item \textbf{Efficient Generative Modeling:} We achieve state-of-the-art performance using standard DiT architectures without the need for computationally expensive architectural changes like width scaling. On the 131M-parameter DiT-B, along with RJF and DINOv2-B achieves an FID of \textbf{3.37} with guidance and FID of 4.95 without guidance in 200 epochs as shown in \cref{fig:intro_grah}, whereas the standard flow matching  fails to converge. These gains persist at scale: on DiT-XL, we attain an FID of \textbf{3.62} in 80 epochs without guidance, outperforming both the standard flow matching (FID 4.28) and the VAE-based DiT trained with alignment losses (FID 4.29).
\end{itemize}

\section{Geometrical Analysis}
\label{sec:motivation}

Following RAE \cite{zheng2025diffusion}, we investigate the feasibility of directly using pretrained representation encoders within the Diffusion Transformer framework and had similar observation that standard diffusion recipe fail to converge effectively, even in a simplified single-image overfitting. Rather than seeking marginal architectural improvements to address this failure, we aim to answer a more fundamental question: \textbf{\textit{Why are these high-dimensional, semantically rich representations resistant to the standard Diffusion recipe?}} To answer this, we first analyze the intrinsic geometry of the feature space generated by these pretrained representation encoders.

\subsection{The Geometry Gap}
\label{sec:Geometry gap}
\begin{wrapfigure}{r}{0.6\textwidth}
    \centering
    \includegraphics[width=1.0\linewidth]{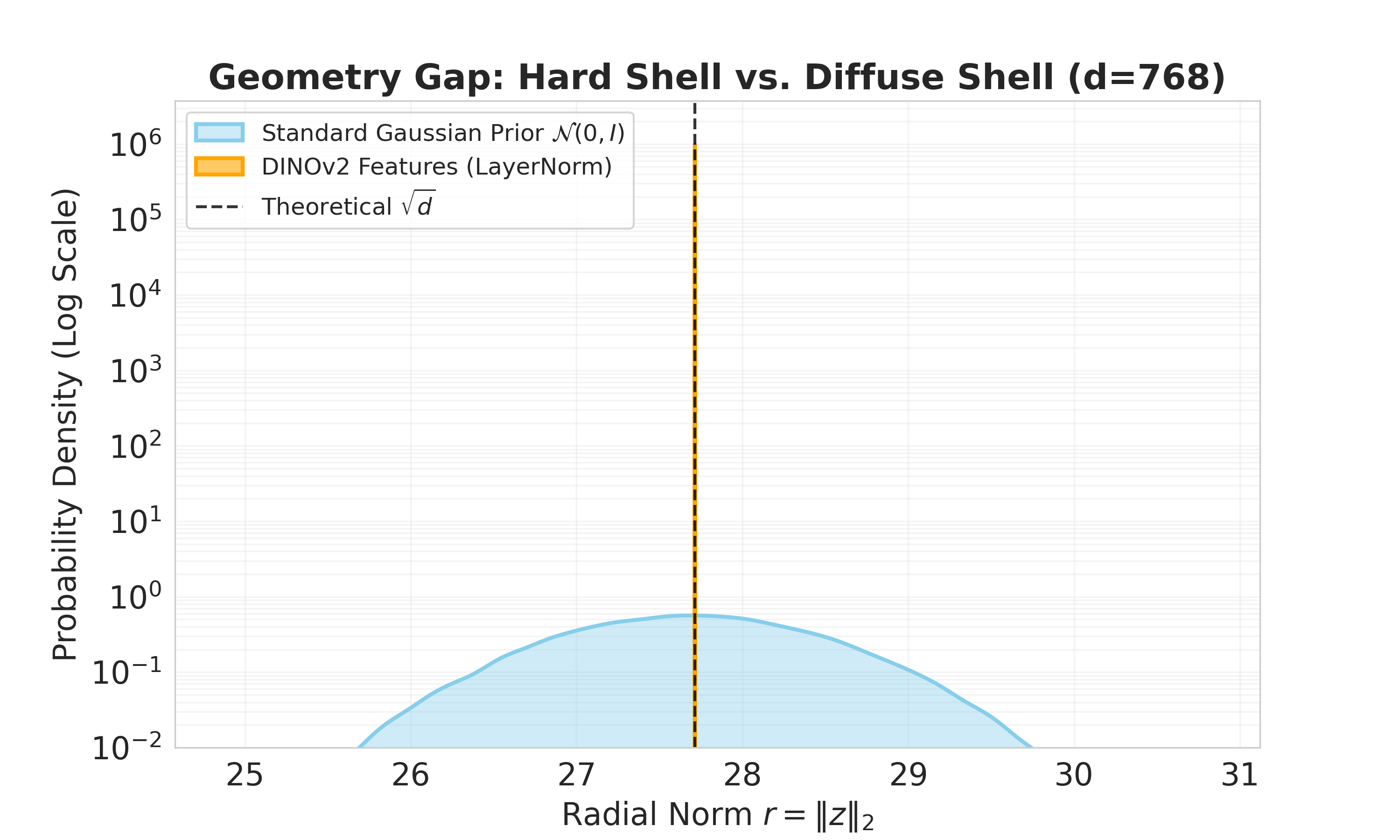}
    \vspace{-10pt}
    \caption{\textbf{The Geometry Gap}. A comparison of radial feature norms ($r = \|z\|_2$) between DINOv2-B representations and a standard Gaussian prior in $\mathbb{R}^{768}$. While the Gaussian prior (blue) is distributed across a diffuse shell, DINOv2-B features (orange) are rigidly constrained to a hypersphere with near-zero radial variance. This extreme geometric mismatch prevents standard diffusion models from converging effectively on representation encoders.}
    \label{fig:thick_shell}
    \vspace{-20pt}
\end{wrapfigure}

We analyze the distribution of the final feature vectors $z \in \mathbb{R}^d$ extracted from the DINOv2-B \cite{oquab2023dinov2} encoder. Decomposing these features into radial and angular components reveals a rigid geometric constraint: the features do not populate the ambient Euclidean space but are explicitly projected onto a hypersphere $\mathcal{S}^{d-1}$ of fixed radius $\sqrt{d}$:
\begin{equation}
    z = r \cdot \hat{z}, \quad \text{where } r \approx \sqrt{d} \text{ and } \hat{z} \in \mathcal{S}^{d-1}.
\end{equation}
As illustrated in Figure \ref{fig:thick_shell}, the radial component $r$ exhibits near-zero variance due to the ubiquitous application of LayerNorm. This creates a hard shell geometry where all semantic information is encoded exclusively in the angular component $\hat{z}$. This stands in sharp contrast to the standard Gaussian prior used in diffusion models, which assumes a probability mass concentrated in a diffuse shell.

This hyperspherical geometry reveals why the standard flow matching become suboptimal. The standard algorithm constructs a conditional probability path $p_t(x)$ via linear interpolation between the source distribution (Gaussian noise $\epsilon$) and the target data ($x$):
\begin{equation}
    x_t = (1 - t)x + t \epsilon.
\end{equation}
In Euclidean space, this linear trajectory is optimal. However, on a hyperspherical manifold, this creates a critical distribution shift as also mentioned in \cite{rozen2021moser, mathieu2020riemannian}. Since $\epsilon$ and $x$ are high-dimensional vectors, they are approximately orthogonal($\epsilon \cdot x \approx 0$). Consequently, the squared norm of the intermediate state $x_t$ follows:
\begin{equation}
    \|x_t\|^2 \approx (1-t)^2 \|x\|^2 + t^2 \|\epsilon\|^2.
\end{equation}

At $t=0.5$, the norm collapses to $\|x_{0.5}\| \approx \frac{1}{\sqrt{2}} \sqrt{d} \approx 0.7 \sqrt{d}$. This implies that the linear flow trajectory $x_t$ does not stay on the manifold $\mathcal{S}^{d-1}$ but rather cuts through the interior of the hypersphere (a chord). This forces the network to learn a velocity field $v_t$ in regions of the feature space that are strictly off manifold for the pretrained representation encoder. The model must essentially hallucinate valid semantic gradients in a region where the representation space is undefined, leading to the convergence failure.

\subsection{Revisiting the Capacity Hypothesis: Geometric Interference}

This convergence failure is also identified in RAE \cite{zheng2025diffusion}, to resolve that they proposed a width scaling solution: increasing the Diffusion Transformer's width ($d_{model}$) to match atleast the token dimension ($n$). Crucially, they demonstrate that this is not just a capacity issue—simply adding layers (depth) fails to improve convergence. They hypothesize that the bottleneck is strictly dimensional: because the added Gaussian noise is full-rank, a model with width $d_{model} < n$ suffers from rank collapse.

While it is true that a narrow model cannot fully resolve high-dimensional Gaussian noise, but rank collapse should not preclude the learning of the data manifold itself, which often lies on a lower-dimensional subspace \cite{pope2021intrinsic}. We hypothesized that the failure is not only due to a lack of capacity to model the signal, but rather \textbf{Geometric Interference}: the standard Euclidean Flow Matching objective forces the model to prioritize a radial error term that conflicts with representation learning.

\begin{figure*}[t]
    \centering
    \includegraphics[width=1.0\linewidth]{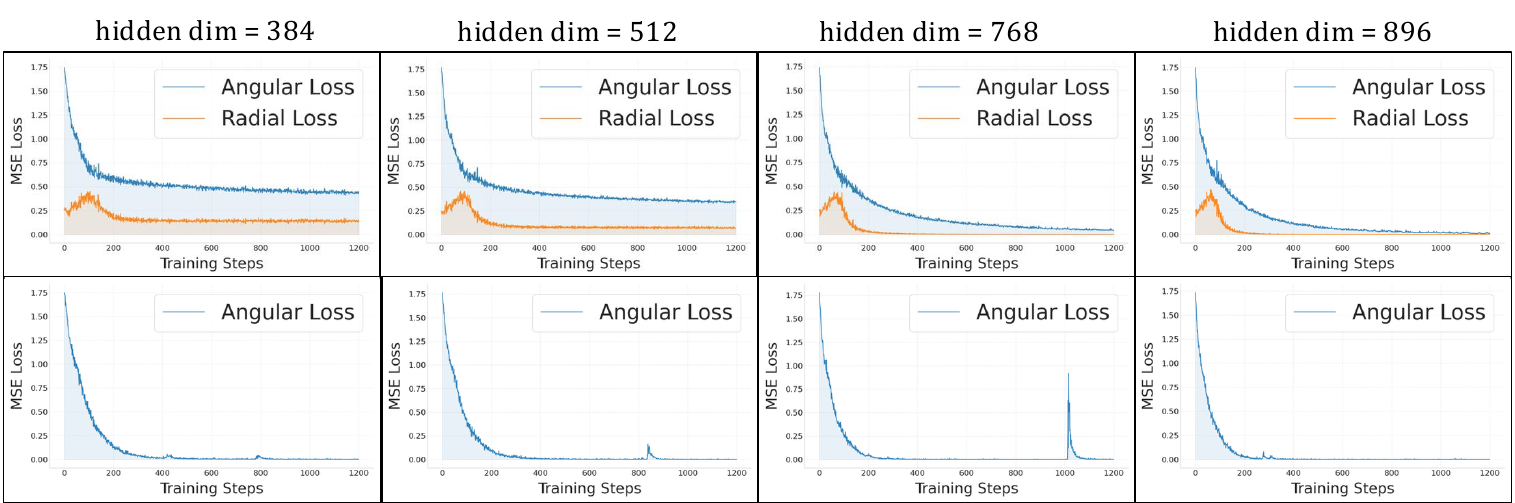} 
    \caption{\textbf{Geometric Interference vs. Capacity.} We train DiT-S models of varying widths on DINOv2 tokens ($d=768$). \textbf{Top Row:} When minimizing Euclidean MSE, narrower models ($d < 768$) suffer from  collapse; the Angular Loss (semantics) gets stuck. 
    \textbf{Bottom Row:} When the radial loss is ignored, even narrow models ($d=384$) converge perfectly on the angular component. This proves the bottleneck is not the dimensionality of the data, but the geometric conflict in the objective.}
    \label{fig:loss_ablation}
\end{figure*}

To test this, we revisited the single-image overfitting setup. We decomposed flow matching loss into radial (magnitude) and angular (direction) components:
\begin{equation}
    \mathcal{L}_{\text{total}} = \underbrace{\| \text{proj}_{\hat{r}}(v_{\text{pred}} - v_{\text{target}}) \|^2}_{\text{Radial Loss }} + \underbrace{\| \text{proj}_{\perp}(v_{\text{pred}} - v_{\text{target}}) \|^2}_{\text{Angular Loss}}.
\end{equation}
We then trained DiT models of varying widths ($d=384$ to $d=896$) on DINOv2-B tokens ($n=768$).

As shown in Figure \ref{fig:loss_ablation}, when optimizing the full Euclidean loss (Top Row), models with width $< n$ (e.g., 384, 512) fail completely. The Angular Loss (blue) —which represents the learning of image semantics—stalls and fails to converge. The model effectively wastes its limited rank trying to minimize the Radial Loss (orange), which arises because the Euclidean interpolation forces a chord trajectory that violates the hyperspherical manifold.

However, when we mask the radial loss and optimize \textit{only} the angular component (Bottom Row), the capacity bottleneck vanishes. Even the smallest model ($d=384$, half the token dimension) converges instantly. This experiment provides a crucial insight: The model has sufficient capacity to learn the semantics, but under Euclidean objective, the radial noise dominates the gradient updates. 

The ``width scaling'' solution proposed by \cite{zheng2025diffusion} is effectively a brute force fix—it grants the model enough parameters to memorize the ill-posed radial vector field through the void. However, we argue that simply masking the radial component or projecting the noise prior onto the manifold is insufficient to resolve this. While these modifications ensure valid endpoints (isolating the angular component), the underlying Euclidean linear trajectory still forms a chord that traverses the manifold's interior, as shown in \cref{sec:Geometry gap}. Instead of scaling the architecture to fit a broken objective, we propose to fix the objective itself. By adopting \textbf{Riemannian Flow Matching} \cite{chen2023flow}, we define the diffusion process directly on the manifold $\mathcal{S}^{d-1}$. This eliminates the radial conflict by design, ensuring that the transport trajectory follows the geodesic (the arc) rather than the chord, naturally aligning the generative process with the representation.

\section{Method}

\subsection{Euclidean Flow Matching}
Flow Matching (FM) \cite{lipman2022flow} is a simulation-free framework for training Continuous Normalizing Flows (CNFs). The goal is to learn a time-dependent vector field $v_t: \mathbb{R}^d \to \mathbb{R}^d$ that generates a probability path $p_t(x)$ transforming a simple prior distribution $\epsilon \sim  \mathcal{N}(0, I)$ to the complex data distribution $x \sim p_{\text{data}}$.

The flow is defined by the ordinary differential equation:
\begin{equation}
    \frac{d x_t}{dt} = v_t(x_t), \quad t \in [0, 1].
\end{equation}
To scale this to high dimensions, Conditional Flow Matching (CFM) trains the model to approximate the conditional vector field generating a specific probability path between a data sample $x \sim p_{\text{data}}$ and noise $\epsilon \sim \mathcal{N}(0, I)$.

In the standard Euclidean setting, the simplest probability path is constructed via linear interpolation (Optimal Transport displacement):
\begin{equation}
    x_t = (1 - t)x + t\epsilon.
\end{equation}
Differentiating with respect to time $t$, the ground-truth conditional velocity field $u_t(x_t | x, \epsilon)$ is constant and straight:
\begin{equation}
    u_t(x_t | x, \epsilon) = \frac{d}{dt}((1 - t)x + t\epsilon) = \epsilon - x.
\end{equation}
The flow matching objective minimizes the mean squared error between the parameterized vector field $v_\theta(x_t, t)$ and the target velocity:
\begin{equation}
    \mathcal{L}_{\text{FM}}(\theta) = \mathbb{E}_{t, p(x), p(\epsilon)} \left[ \| v_\theta(x_t, t) - (\epsilon - x) \|^2 \right].
\end{equation}

\subsection{Riemannian Flow Matching on Hyperspherical Manifolds}
\label{sec:riemannian_fm}

While Euclidean Flow Matching has driven recent advances in latent generative modeling, our analysis in Section \ref{sec:motivation} demonstrates that it is fundamentally ill-suited for the hyperspherical feature spaces produced by representation encoders. The standard linear interpolant violates the manifold structure, forcing the model to learn a vector field through the undefined interior of the sphere. 

To resolve this, we propose to reformulate the diffusion process directly on the intrinsic data manifold. 
We first project our feature vectors to the unit norm hypersphere $\mathcal{M} = \mathcal{S}^{d-1} \subset \mathbb{R}^d$ and define our source distribution by projecting isotropic Gaussian noise onto the manifold $\mathcal{M}$.

\textbf{Geodesic Probability Paths}
In the Euclidean setting, the optimal transport path between a source $x$ and target $\epsilon$ is a straight line (a chord). On the hypersphere $\mathcal{S}^{d-1}$, the optimal path is the geodesic.

The conditional probability path $x_t$ is defined via Spherical Linear Interpolation (SLERP) rather than linear interpolation. Given data $x \in \mathcal{S}^{d-1}$ and noise $\epsilon \in \mathcal{S}^{d-1}$ (where $\|\epsilon\|=1$), the geodesic path is given by:
\begin{equation}
    x_t = \text{SLERP}(x, \epsilon; t) = \frac{\sin((1-t)\Omega)}{\sin(\Omega)} x + \frac{\sin(t\Omega)}{\sin(\Omega)} \epsilon
\end{equation}
where $\Omega = \arccos(x^\top \epsilon)$ is the geodesic distance (angle) between the data and the noise. Unlike the Euclidean path, this trajectory ensures that $\|x_t\| = 1$ for all $t \in [0, 1]$, completely eliminating the norm collapse phenomenon and ensuring the generative process on representation manifold.

\textbf{Tangent Space Velocity Fields}
A critical consequence of restricting the flow to $\mathcal{M}$ is that the velocity vector $v_t$ must essentially lie in the tangent space $\mathcal{T}_{x_t}\mathcal{M}$ at every point $x_t$. For the sphere, this implies the velocity must be orthogonal to the position vector: $v_t \cdot x_t = 0$.

The target Riemannian velocity field $u_t^{\mathcal{M}}(x_t | x, \epsilon)$ is computed by differentiating the geodesic path with respect to time $t$:
\begin{equation}
    \begin{split}
        u_t^{\mathcal{M}}(x_t) &= \frac{d}{dt} \text{SLERP}(x, \epsilon; t) \\
        &= \frac{\Omega}{\sin(\Omega)} \Big( \cos(t\Omega)\epsilon - \cos\big((1-t)\Omega\big)x \Big).
    \end{split}
    \label{eq:riemannian_velocity}
\end{equation}

\textbf{The Riemannian Objective}
Consequently, we replace the standard objective with the Riemannian Flow Matching loss. We train the network $v_\theta$ to predict this tangent vector field. Crucially, since the target $u_t^{\mathcal{M}}$ lies strictly in the tangent space, the radial component of the error is structurally zero by design. The loss simplifies to the squared norm in the ambient space, which is equivalent to the Riemannian metric induced on the sphere:
\begin{equation}
    \mathcal{L}_{\text{RFM}}(\theta) = \mathbb{E}_{t, p(x), p(\epsilon)} \left[ \| v_\theta(x_t, t) - u_t^{\mathcal{M}}(x_t) \|^2 \right].
\end{equation}
By optimizing this objective, the model learns purely semantic transitions (angular changes), effectively resolving the Geometric Interference identified in \cref{sec:motivation}.

To preserve the constant-speed advantage during sampling, we use  Geodesic (Exponential Map) Integration. Rather than moving along a straight tangent line that drifts off the manifold, the exponential map wraps the velocity vector around the sphere's surface. 

For a point $x_t \in \mathcal{S}^{d-1}$ and a predicted tangent velocity $v \in \mathcal{T}_{x_t}\mathcal{S}^{d-1}$ , the update is defined by the closed-form trigonometric rotation:
\begin{equation}
x_{t+\Delta t} = \cos(\|v\|\Delta t)x_t + \sin(\|v\|\Delta t)\frac{v}{\|v\|}.
\end{equation}

This update ensures the trajectory follows the great circle exactly, matching the Riemannian flow learned during training. To correct for minor numerical drift over many integration steps, we perform a final rotate and normalize operation. This approach provides a computationally efficient, simulation-free inference path that maintains the rigid DINO geometry without the distortion artifacts of Euclidean solvers.

\subsection{Jacobi Field Regularization}\label{sec:jacobi}
While Riemannian Flow Matching with SLERP ensures that the generated path stays on the manifold, the standard velocity-matching objective remains geometrically unaware. The loss $\mathcal{L}_{\text{RFM}} = \|v_\theta - u_t\|^2$ implicitly assumes a flat metric, treating velocity errors uniformly across time $t \in [0, 1]$. However, on a positively curved manifold $\mathcal{S}^{d-1}$, the impact of a velocity error is not uniform. Due to the focusing of geodesics, a perturbation in the velocity vector $w \in \mathcal{T}_{x_t}\mathcal{M}$ propagates non-linearly. To maximize generation fidelity in high dimensional space, we must prioritize minimizing the error near the noise (the endpoint $\epsilon$, $t=1$).

Inspired by \cite{zaghen2025riemannian}, we model this error propagation using Jacobi Fields, which quantify the separation between geodesics caused by velocity perturbations. Solving the Jacobi equation for a hypersphere yields a geometric weighting factor $\lambda(t, \Omega)$ that scales the loss based on the curvature-induced focusing of geodesics:
\begin{equation}
    \lambda(t, \Omega) = \text{sinc}^2((1-t)\Omega),
\end{equation}
where $\Omega$ is the total geodesic distance. This term acts as a geometry-aware attention mechanism: it down-weights errors near $t=0$ (Data) where geodesic focusing mitigates perturbations, and prioritizes precision near $t=1$ (noise) where the generative trajectory must precisely align with the feature manifold. The final Jacobi-Regularized objective is:
\begin{equation}
    \mathcal{L}_{\text{Jacobi}}(\theta) = \mathbb{E}_{t, x, \epsilon} \left[ \lambda(t, \Omega) \cdot \| v_\theta(x_t, t) - u_t^{\mathcal{M}}(x_t) \|^2 \right].
\end{equation}
By optimizing this curvature-corrected objective, we effectively anneal the learning signal, forcing the model to prioritize the learning of high-dimensional latent space. Further details are provided in supplementary.


\section{Experiments}

\subsection{Implementation Details}
To ensure a fair comparison, we follow the training protocol of LightingDiT \cite{yao2025reconstruction}. Experiments are conducted on ImageNet-1K \cite{russakovsky2015imagenet} at $256 \times 256$ resolution. Unless otherwise specified, we use LightingDiT \cite{yao2025reconstruction} as our base architecture and train for 80 epochs with a global batch size of 1024. We use the RAE decoder \cite{zheng2025diffusion} for all representation encoders. Training utilizes the Adam optimizer ($\beta_1 = 0.9, \beta_2 = 0.95$) with a fixed learning rate of $2 \times 10^{-4}$. We apply gradient clipping with a maximum norm of $1.0$ and maintain an Exponential Moving Average (EMA) of weights with a decay of $0.9995$. We follow the same setting of dimension dependent noise schedule shift of RAE \cite{zheng2025diffusion} with n=4096. We use Autoguidance \cite{karras2024guiding} as our guidance method with guidance scale of 1.6. For inference, we use an Geodesic (Exponential Map) integrator with 50 steps and evaluate performance on 50k generated images.

\begin{wraptable}{r}{0.5\textwidth}
    \centering
    \begin{minipage}{\linewidth} 
        \footnotesize
        \vspace{-13pt}
\caption{\textbf{FID comparison on ImageNet 256$\times$256 without guidance} across various model sizes for LightingDiT with REPA, DiNOv2-B with Euclidean Flow matching (EFM) and RJF.}
\label{tab:Convergence}
\begin{adjustbox}{max width=\linewidth}
\begin{tabular}{l c c c}
\toprule
Model & \#Params & Epochs. & FID$\downarrow$ \\
\midrule
DiT-B/2       & 130M & 80 & 43.47 \\
LightningDiT-B/1 & 130M & 80 & 22.86 \\
+~REPA               & 130M & 80 & 21.45 \\
+~EFM (DiNOv2-B)  & 131M & 80 & 24.21 \\
\cellcolor{cyan!15} +~RJF (DiNOv2-B) (Ours)  & \cellcolor{cyan!15} 131M & \cellcolor{cyan!15} 80 & \cellcolor{cyan!15} \textbf{6.77} \\
\midrule
DiT-L/2       & 458M & 80 & 23.33 \\
LightningDiT-L/1 & 458M & 80 & 10.08 \\
+~REPA               & 458M & 80 & 7.48 \\
+~EFM (DiNOv2-B)  & 459M & 80 & 6.31\\
\cellcolor{cyan!15} +~RJF (DiNOv2-B) (Ours)  & \cellcolor{cyan!15} 459M & \cellcolor{cyan!15} 80 & \cellcolor{cyan!15} \textbf{4.21} \\
\midrule
DiT-XL/2     & 675M & 80 & 19.47 \\
LightningDiT-XL/1 & 675M & 80 & 9.29 \\
+~REPA               & 675M & 80 & 6.94 \\

+~EFM (DiNOv2-B)  & 677M & 14 & 10.23 \\
+~EFM (DiNOv2-B)  & 677M & 24 & 7.93 \\
+~EFM (DiNOv2-B)  & 677M & 80 & 4.28 \\
\cellcolor{cyan!15} +~RJF (DiNOv2-B) (Ours)   & \cellcolor{cyan!15} 677M & \cellcolor{cyan!15} 14 & \cellcolor{cyan!15}8.83 \\
\cellcolor{cyan!15}+~RJF (DiNOv2-B) (Ours)  & \cellcolor{cyan!15}677M & \cellcolor{cyan!15}24 & \cellcolor{cyan!15} 6.32 \\
\cellcolor{cyan!15}+~RJF (DiNOv2-B) (Ours)   & \cellcolor{cyan!15}677M & \cellcolor{cyan!15}80 & \cellcolor{cyan!15} \textbf{3.62} \\
\bottomrule
\end{tabular}
\end{adjustbox}
\vspace{-10pt}
    \end{minipage}
\end{wraptable}

\subsection{Main Results}

\textbf{Scaling and Training Convergence.} 
We evaluate the convergence and scalability of our method on ImageNet $256 \times 256$ generation without guidance, comparing it against DiT~\cite{peebles2023scalable}, LightingDiT~\cite{yao2025reconstruction}, REPA~\cite{yu2024representation}, and a baseline using DiNOv2-B features with Euclidean Flow Matching (EFM) as shown in \cref{tab:Convergence}. Our method consistently achieves superior FID performance while significantly accelerating convergence across all evaluated model scales. For the DiT-B architecture trained for 80 epochs, our method reduces FID from 21.45 (REPA) and 24.21 (EFM) to 6.77, demonstrating the critical importance of respecting the underlying feature geometry. In the DiT-L setting, we observe a similar trend, where our approach reduces FID from 10.08 to 4.21 compared to LightningDiT. Notably, in the large-scale setting (DiT-XL), our method demonstrates superior convergence efficiency; at just 24 epochs, it achieves an FID of 6.32, outperforming the strong REPA baseline trained for the full 80 epochs (6.94). By 80 epochs, our method reaches an FID of 3.62, outperforming the Euclidean baseline  by 1.19.

\textbf{State-of-Art Comparison.} Due to computational constraints, we benchmark our method in the limited 80-epoch training regime (Table \ref{tab:sota-comparison}). Our LightingDiT -XL model trained with RJF achieves a highly competitive FID of \textbf{3.62}, significantly outperforming the Euclidean Flow Matching baseline (FID 4.28) trained on DINOv2-B features. Crucially, our method demonstrates superior semantic fidelity compared to all other methods. We achieve a state-of-the-art IS of 186.2 and Precision of 0.82, surpassing recent sota methods. This indicates that while our geometric alignment improves FID, it particularly excels at capturing the high-fidelity semantic modes of the data distribution.

\begin{table*}[h]
\centering
\small
\renewcommand{\arraystretch}{0.9}
\setlength{\tabcolsep}{8pt} 
\caption{\textbf{Class-conditional performance on ImageNet 256$\times$256 with and without guidance.} Our method achieves a superior FID of \textbf{3.62}, outperforming the standard flow matching baseline (FID 4.28).}
\label{tab:sota-comparison}
\label{tab:comparison_perf}
\makebox[\textwidth][c]{
\resizebox{0.9\textwidth}{!}{
\begin{tabular}{l c c cccc cccc}
\toprule
\multirow{2}{*}{\textbf{Method}} & \multirow{2}{*}{\makecell{\textbf{Epochs}}} & \multirow{2}{*}{\textbf{\#Params}} & \multicolumn{4}{c}{\textbf{Generation@256 w/o guidance}} & \multicolumn{4}{c}{\textbf{Generation@256 w/ guidance}} \\
\cmidrule(lr){4-7} \cmidrule(lr){8-11}
 & & & \textbf{FID}$\downarrow$ & \textbf{IS}$\uparrow$ & \textbf{Prec.}$\uparrow$ & \textbf{Rec.}$\uparrow$ & \textbf{FID}$\downarrow$ & \textbf{IS}$\uparrow$ & \textbf{Prec.}$\uparrow$ & \textbf{Rec.}$\uparrow$ \\
\arrayrulecolor{black}\midrule
\multicolumn{11}{l}{\textit{\textbf{Pixel Diffusion}}} \\
\arrayrulecolor{black!30}\midrule
ADM~\cite{dhariwal2021diffusion} &  400  &  554M  & 10.94 &  101.0 & 0.69 & 0.63 & 3.94 & 215.8 & \textbf{0.83} & 0.53\\
RIN~\cite{jabri2022scalable} &  480  &  410M  & 3.42  & 182.0  &  -   & -&  -   &  -      &  -   &  -  \\
PixelFlow~\cite{chen2025pixelflow} & 320 & 677M & -  & -  &   -  &  -   & 1.98 & 282.1 & 0.81 & 0.60 \\
PixNerd~\cite{wang2025pixnerd} & 160 & 700M & -    & -      &  - &  -  &  2.15 & 297.0 & 0.79 & 0.59 \\
SiD2~\cite{hoogeboom2024simpler} &  1280   &   - & -   &  - &  -   &  -   &  1.38    & -   &  -   &  - \\
\arrayrulecolor{black}\midrule
\multicolumn{11}{l}{\textit{\textbf{Vanilla Latent Diffusion}}} \\
\arrayrulecolor{black!30}\midrule
DiT~\cite{peebles2023scalable} & 1400 & 675M         & 9.62 & 121.5 & 0.67 & 0.67 & 2.27 & 278.2 & \textbf{0.83} & 0.57 \\
MaskDiT~\cite{zheng2023fast} & 1600 & 675M & 5.69 & 177.9 & 0.74 & 0.60 & 2.28 & 276.6 & 0.80 & 0.61 \\
SiT~\cite{ma2024sit} & 1400 & 675M         & 8.61 & 131.7 & 0.68 & 0.67 & 2.06 & 270.3 & 0.82 & 0.59 \\
TREAD~\cite{krause2025tread} & 740 & 675M & - & - & - & - & 1.69 & 292.7 & 0.81 & 0.63 \\
MDTv2~\cite{gao2023mdtv2} & 1080 & 675M & -&  - & - & - & 1.58 & 314.7 & 0.79 & 0.65 \\

\arrayrulecolor{black}\midrule
\multicolumn{11}{l}{\textit{\textbf{Latent Diffusion with Self-supervised Representation Model}}} \\
\arrayrulecolor{black!30}
\midrule

REPA~\cite{yu2024representation} & 800 & 675M & 5.90 & 157.8 & 0.70 & \cellcolor{orange!6}  \textbf{0.69} & 4.70 & 305.7 & \cellcolor{orange!6} \textbf{0.80} & 0.65 \\
REPA-E~\cite{leng2025repa} & 800 & 675M & 1.83 & 217.3 & - & - & 1.26 & \cellcolor{orange!6} \textbf{314.9} & 0.79 & 0.66 \\
REG~\cite{wu2025representation} & 480 & 677M & 2.20 & 219.1 & 0.77 & 0.66 & 1.40 & 296.9 & 0.77 & 0.66 \\
LightningDiT~\cite{yao2025reconstruction} & 800 & 675M & 2.17 & 205.6 & 0.77 & 0.65 & 1.35 & 295.3 & 0.79 & 0.65 \\
DDT~\cite{wang2025ddt} & 800 & 675M & 6.27 & 154.7 & 0.68 & \cellcolor{orange!6}  \textbf{0.69} & 1.26 & 310.6 & 0.79 & 0.65 \\
RAE (DiT$^{\text{DH}})$~\cite{zheng2025diffusion} & 800 & 839M & \cellcolor{orange!6} \textbf{1.60} & \cellcolor{orange!6}\textbf{242.7} & \cellcolor{orange!6} \textbf{0.79} & 0.65 & 1.28 & 262.9 & 0.78 & \cellcolor{orange!6} \textbf{0.67}\\
SFD (XL)~\cite{pan2025semantics} & 800 & 676M & 2.54  & - & - & -& \cellcolor{orange!6} \textbf{1.06} & 267.0 & 0.78 & \cellcolor{orange!6} \textbf{0.67} \\
\arrayrulecolor{black}\midrule

REPA~\cite{yu2024representation} & 80& 675M & 7.90 & 122.6 & 0.70 & 0.65 & - & - & - & - \\
REPA-E~\cite{leng2025repa} & 80 & 675M & 3.46 & 159.8 & 0.77 & 0.63 & 1.67 & 266.3 & 0.80 & 0.63 \\
REG~\cite{wu2025representation} & 80 & 677M & \cellcolor{purple!25} \textbf{3.40} & 184.1 & 0.77 & 0.63 & 1.86 & \cellcolor{purple!25}\textbf{321.4} & 0.76 & 0.63  \\
LightningDiT~\cite{yao2025reconstruction} & 64 & 675M & 5.14 & 130.2 & 0.76 & 0.62 & 2.11 & 252.3 & 0.81 & 0.58 \\
SVG~\cite{shi2025latent} & 80 & 675M & 6.57 & 137.9 & - & - & 3.54 & 207.6 & - & - \\
SFD (XL)~\cite{pan2025semantics} & 80 &  675M & 3.53  & 162.0 & 0.75 & \cellcolor{purple!25} \textbf{0.65} & \cellcolor{purple!25} \textbf{1.30} & 233.4 & 0.78 & \cellcolor{purple!25}\textbf{0.64} \\
DiT-XL(DiNOv2-B)~\cite{yao2025reconstruction} & 80 & 677M & 4.28 & - & - & - & -& - & - & - \\
DiT-XL(DiNOv2-B) + RJF (Ours) & 80 & 677M & 3.62  & \cellcolor{purple!25} \textbf{186.2} & \cellcolor{purple!25} \textbf{0.82} & 0.52 & 2.81 & 201.22 & \cellcolor{purple!25} \textbf{0.82} & 0.56 \\

\arrayrulecolor{black}\bottomrule
\end{tabular}
}
}
\end{table*}

\begin{wrapfigure}{r}{0.5\textwidth}
    \centering
     \vspace{-25pt}
    \includegraphics[width=1.0\linewidth]{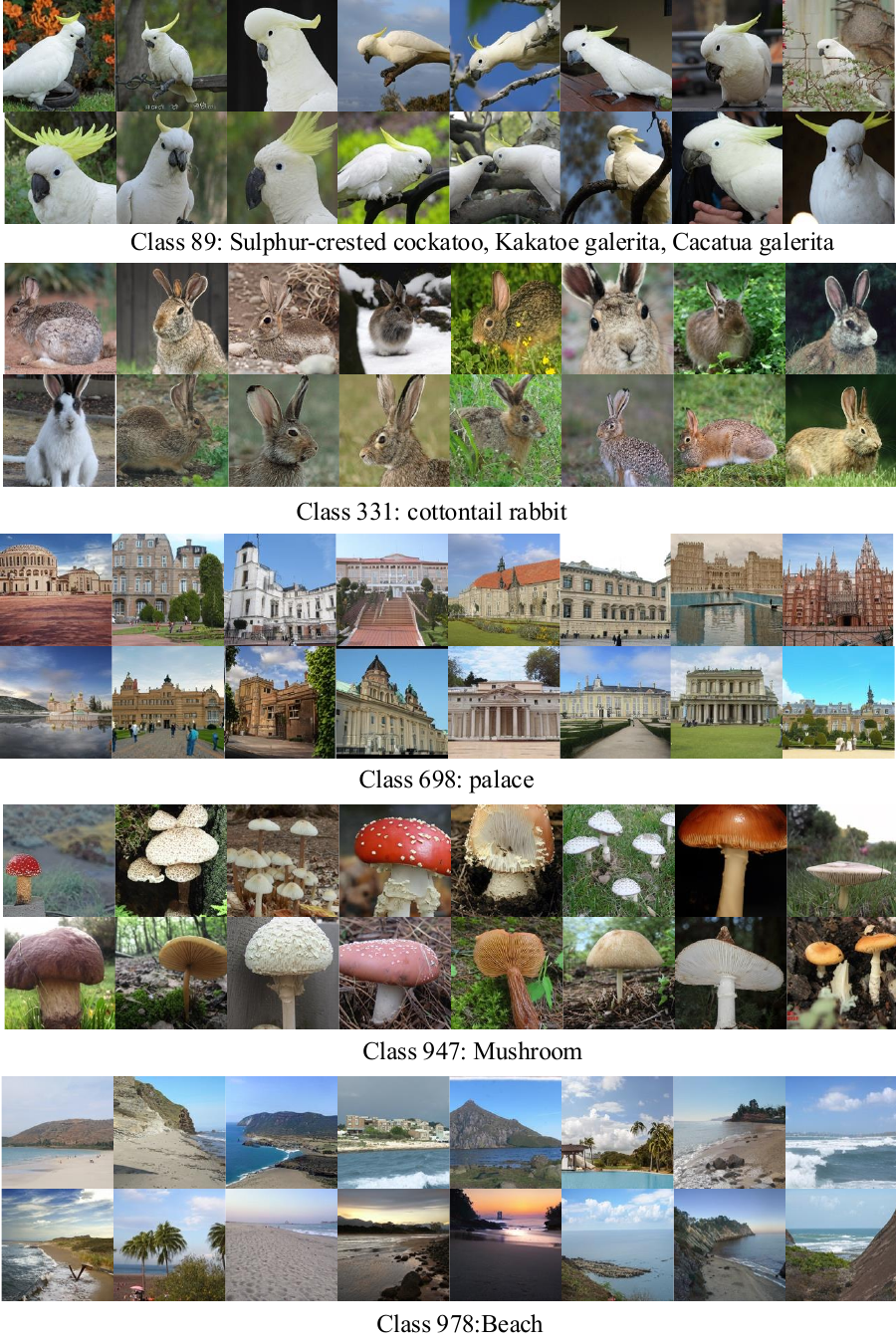}
    \vspace{-20pt}
    \caption{\textbf{Qualitative results of LightingDiT-XL+RJF trained for 80 epochs on ImageNet 256$\times$256.} We show uncurated results on the five classes .}
    \label{fig:qualitative_results}
    \vspace{-20pt}
\end{wrapfigure}

In \cref{fig:qualitative_results}, we present uncurated qualitative samples from our LightingDiT-XL model trained with RJF on ImageNet $256\times256$. Notably, the model achieves high generation quality and semantic diversity after only 80 epochs of training. More uncurated qualitative results are provided in Supplementary. 

\subsection{Ablation Study}
\label{sec:ablation}

We investigate the impact of each geometric component by training a standard LightingDiT-B/1 model on DINOv2-B latents. The results are summarized in Table \ref{tab:ablation_study}. When trained with standard Euclidean Flow Matching(EFM), the model fails to converge effectively, yielding a poor FID of \textbf{24.32}. As analyzed in \cref{sec:motivation}, baseline suffers from Geometric Interference: the model wastes capacity minimizing radial errors and learning velocity fields within the undefined interior of the manifold.


\begin{wraptable}{r}{0.6\textwidth}
    \centering
    \begin{minipage}{\linewidth} 
        \footnotesize
        \vspace{-40pt}
\setlength{\tabcolsep}{7pt}
\caption{\textbf{Ablation of Geometric Components.} We train a LightingDiT-B model on DINOv2-B features. The Standard Euclidean baseline fails to converge (FID   24.32) due to geometric interference. Projecting noise to sphere (\textbf{+SN}) yields only marginal gains, as the linear path remains flawed. Adopting \textbf{Riemannian Flow Matching (+RFM)} resolves the trajectory mismatch, drastically improving FID to 7.06, with \textbf{Jacobi Regularization (+RJF)}, achieves SOTA performance (FID \textbf{6.77}), demonstrating that respecting geometry eliminates need for width scaling, further training leads to FID $\textbf{3.37}$.}
\label{tab:ablation_study}
\begin{adjustbox}{max width=0.95\linewidth}
\begin{tabular}{lcrrrr}
\toprule
\textbf{ Method } 
 & Epochs & FID$\downarrow$  & IS$\uparrow$ & Prec.$\uparrow$ & Rec.$\uparrow$ \\
\midrule
DiT-B/1(DiNOv2-B)  & 80 & 24.32  & 79.34 & 0.63 & 0.46 \\
DiT-B/1(DiNOv2-B)  + SN & 80  & 21.99  & 98.25 & 0.62 & 0.47 \\
DiT-B/1(DiNOv2-B)  + RFM  & 80  & 7.06  & 136.70 & 0.78 & 0.49 \\
\arrayrulecolor{black}\midrule
DiT-B/1(DiNOv2-B)  + RJF &   80 & 6.77 & 138.12 & 0.78 & 0.50 \\
\cellcolor{cyan!15}   DiT-B/1(DiNOv2-B)  + RJF & \cellcolor{cyan!15}  200  & \cellcolor{cyan!15}  \textbf{4.95}  & \cellcolor{cyan!15}  \textbf{157.48} & \cellcolor{cyan!15}  \textbf{0.79} & \cellcolor{cyan!15} \textbf{0.52} \\
\cellcolor{cyan!15}   DiT-B/1(DiNOv2-B)  + RJF w/ guid & \cellcolor{cyan!15}  200  & \cellcolor{cyan!15}  \textbf{3.37}  & \cellcolor{cyan!15}  \textbf{180.26} & \cellcolor{cyan!15}  \textbf{0.80} & \cellcolor{cyan!15} \textbf{0.56} \\
 
\arrayrulecolor{black}\bottomrule
\vspace{-45pt}
\end{tabular}
\end{adjustbox}
    \end{minipage}
\end{wraptable}

To address the radial error, we project both the Gaussian noise and the target latents onto the unit sphere, effectively forcing the model to learn only the angular component. We observe only a marginal improvement of 2.33 FID (reaching \textbf{21.99}). This indicates that simply fixing the endpoints is insufficient; the issue persists because the Euclidean linear interpolation still forms a chord that violates the manifold geometry.

By adopting Riemannian Flow Matching to ensure the generative trajectory follows the geodesic, we observe a huge performance improvement. The FID drops significantly to \textbf{7.06} (with IS improving to 136.70). This result is critical: it validates our claim that respecting intrinsic geometry enables standard Diffusion Transformers to model high-dimensional features without the need for architectural width scaling. Finally, incorporating Jacobi Regularization to weight the loss according to geodesic focusing further refines the geometric alignment, improving FID to \textbf{6.77}. When trained for 200 epochs, our method (RJF) achieves a remarkable FID of \textbf{4.95}. With guidance, the performance reaches SOTA with an FID of \textbf{3.37} and an IS of \textbf{180.26}.


\begin{wraptable}{r}{0.4\textwidth}
    \centering
    \vspace{-40pt}
    \begin{minipage}{\linewidth} 
        \footnotesize
        \caption{\textbf{Performance with different architecture design with DiNOV2-B }}
        \label{tab:ablation_arch}
        \centering
        \begin{adjustbox}{max width=1.0\linewidth}
        \begin{tabular}{lr}
            \toprule
            \textbf{Method} & FID$\downarrow$ \\
            \midrule
            DiT-XL/1 & 4.29 \\
            DiT-XL/1 & 4.28 \\
            \rowcolor{cyan!15} DiT-XL/1 + RJF (ours) & \textbf{3.62} \\
            \arrayrulecolor{black}\midrule
            DDT-XL/2 & 6.62 \\
            DDT-XL/1 & 6.55 \\
            \rowcolor{cyan!15} DDT-XL/1 + RJF (ours) & \textbf{5.82} \\
            \arrayrulecolor{black}\midrule
            $DiT^{DH}$ & 6.33 \\
            \rowcolor{cyan!15}$DiT^{DH}$ + RJF (ours) & \textbf{6.20} \\
            \bottomrule
        \end{tabular}
        \end{adjustbox}
        \vspace{-20pt}
    \end{minipage}
\end{wraptable}

\section{Discussion}

\subsection{Generality Across Architectures}

To validate the robustness of our approach, we evaluate RJF across diverse Diffusion Transformer architectures (Table \ref{tab:ablation_arch}). 

 We observe consistent performance gains in all settings. On the large-scale LightingDiT-XL, our method achieves an FID of \textbf{3.62}, significantly outperforming both the standard VAE-based baseline (FID 4.29) and the Euclidean Flow Matching on DINOv2 features (FID 4.28). This confirms that our gains scale effectively to larger models. We observe similar improvements on the DDT-XL architecture, where RJF lowers the FID to \textbf{5.82}, surpassing the Euclidean baseline (FID 6.55). We also evaluate $DiT^{DH}-S$, an architecture explicitly designed with width scaling to handle RAE latents. Even in this specialized setting, our geometric objective yields further improvements (reaching FID \textbf{6.20}).

\subsection{Different representation encoders}

\begin{wraptable}{r}{0.4\textwidth} 
    \centering
    \vspace{-50pt}
    \begin{minipage}{\linewidth} 
        \centering
        \caption{\textbf{Ablation study with different Representation encoder}}
        \label{tab:ablation_encoder}
        \begin{adjustbox}{max width=\linewidth}
        \begin{tabular}{lcc}
            \toprule
            & \multicolumn{2}{c}{FID ($\downarrow$)} \\
            \cmidrule(lr){2-3}
            Method & SIGLIP & MAE \\
            \midrule
            DiT-B/1       & 130.21 & 50.48 \\
            \cellcolor{cyan!15} DiT-B/1 + RJF & \cellcolor{cyan!15} \textbf{10.39} & \cellcolor{cyan!15} \textbf{19.82} \\
            \bottomrule
        \end{tabular}
        \end{adjustbox}
        \vspace{-20pt}
    \end{minipage}
\end{wraptable}

To assess the generalization of our method beyond DINOv2, we evaluate RJF on two distinct representation encoders: SigLIP (contrastive) and MAE (reconstructive). As shown in Table \ref{tab:ablation_encoder}, the standard LightingDiT-B/1 baseline fails to converge effectively on either representation, yielding poor FIDs of \textbf{130.21} for SigLIP and \textbf{50.48} for MAE.

We attribute this failure to shared geometric properties of these latent spaces. SigLIP is trained with a contrastive objective that explicitly enforces hyperspherical normalization \cite{wang2020understanding}. Similarly, while MAE is reconstructive, the extensive application of LayerNorm strictly constrains its features to a hyperspherical manifold. Consequently, our method respects intrinsic geometry resolves the convergence issues in both cases. RJF achieves a dramatic improvement, reaching an FID of \textbf{10.39} on SigLIP and \textbf{19.82} on MAE, demonstrating that geometric alignment is essential for learning on representation feature spaces.

\subsection{Projection with different radius}

\begin{wrapfigure}{r}{0.35\textwidth} 
    \centering
    \vspace{-25pt}
    \includegraphics[width=\linewidth]{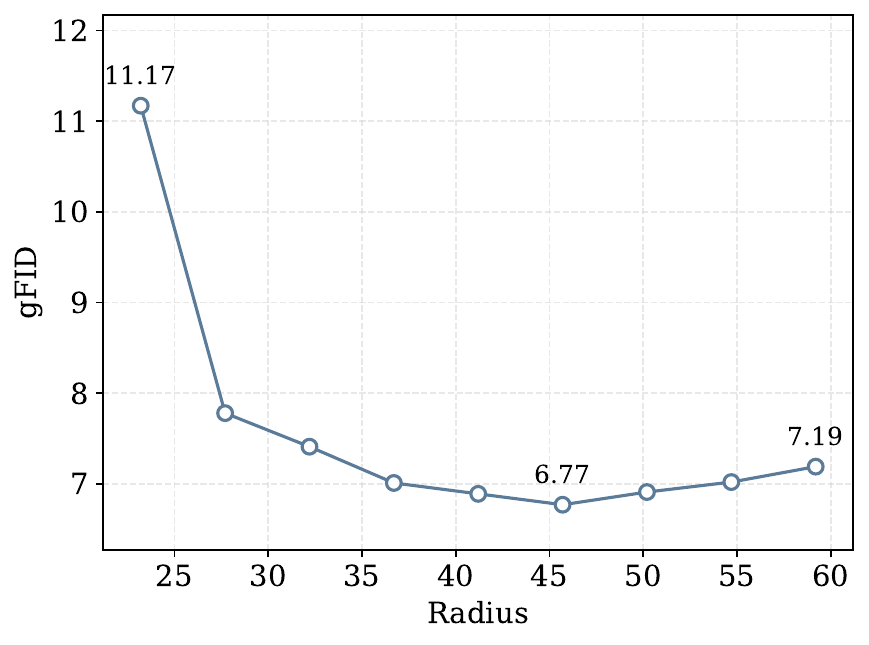}
    \vspace{-20pt}
    \caption{\textbf{Performance across different radius projection}}
    \label{fig:radius}
    \vspace{-20pt} 
\end{wrapfigure}

We further analyze the impact of the projection radius $R$ during the inference stage  (\cref{fig:radius}). While the model is trained on the intrinsic geometry, we observe that strictly reprojecting the generated latents back to the original DINOv2-B norm ($R \approx 27.7$) yields a suboptimal FID of 7.79. Interestingly, increasing the 
projection radius during inference leads to consistent performance gains, achieving the best FID of \textbf{6.77} at a radius of $R \approx 45$. This observation suggests that the RAE decoder is sensitive to feature magnitude; amplifying the norm of the generated latents leading to enhance the fidelity of the generated images.

\section{Related Works}
\label{related_works}

\subsection{Representation Alignment for image generation}

 Recent research has increasingly focused on bridging the gap between generative models and pretrained representation encoders to enhance diffusion transformer performance. A foundational approach in this domain involves feature-space alignment, where methods like REPA \cite{yu2024representation} accelerate convergence by aligning intermediate diffusion features with pretrained representations such as DINOv2 \cite{oquab2023dinov2}. This paradigm has been extended by architectures like DDT \cite{wang2025ddt}, which applies alignment to a decoupled encoder-decoder structure, and REG \cite{wu2025representation}, which introduces a learnable class token for explicit semantic guidance. Beyond feature alignment, significant efforts have been made to optimize the latent space itself; REPA-E \cite{leng2025repa} enables joint optimization of the VAE and diffusion model, ReDi \cite{kouzelis2025boosting} jointly learns low-level and high-level semantic distributions, and approaches like VA-VAE \cite{yao2025reconstruction}, and SVG \cite{shi2025latent}  enrich the conventional VAE with pretrained visual encoder representations. Most recently, RAE \cite{zheng2025diffusion} proposes replacing the VAE entirely with representation encoders. However, to mitigate the training convergence failure associated with this approach, RAE relies on width scaling. While we similarly observe that standard diffusion recipes fail in this setting, we attribute this not due to model capacity, but to a fundamental geometric mismatch. We demonstrate that standard Euclidean flow matching trajectories inadvertently traverse the low-density interior of the feature hypersphere, necessitating a geometrically consistent modeling approach rather than incremental architectural modifications.

\subsection{Riemannian Flow Matching.}
While diffusion models rely on stochastic differential equations, Flow Matching (FM) \cite{lipman2022flow} has emerged as a robust, simulation-free alternative for training Continuous Normalizing Flows (CNFs). Standard FM constructs probability paths via linear interpolation in Euclidean space, regressing a velocity field to guide samples from a source distribution to the data. However, as noted in recent geometric deep learning literature \cite{rozen2021moser, mathieu2020riemannian}, Euclidean linear paths are ill-suited for data residing on non-Euclidean manifolds, as they violate the intrinsic geometry of the domain. To address this, Riemannian Flow Matching (RFM) \cite{chen2023flow} generalizes the framework by replacing Euclidean straight lines with geodesic paths defined by the Riemannian metric. By defining the conditional probability path as a geodesic interpolation, RFM ensures that the flow remains strictly on the manifold, avoiding regions of low density such as the interior of a hypersphere. This formulation has been widely adopted for generative modeling on structured scientific domains, including protein backbone generation on $SE(3)$ \cite{bose2023se, yim2023fast}, torsion angle prediction on tori \cite{jing2022torsional}, and motion planning for robotics on configuration manifolds \cite{braun2024riemannian}.

\section{Conclusion}

In this work, we demonstrate that the failure of standard diffusion transformer on representations encoder(like DiNO, SigLip and MAE) is not a capacity issue, but geometric. We identified that standard Euclidean objectives suffer from Geometric Interference, wasting computation on the minimizing radial error and learning trajectories through the off-manifold area induced
by this geometric mismatch. We solved this with Riemannian Flow Matching with Jacobi Regularization (RJF), which enforces geodesic trajectories consistent with the latent topology. Crucially, RJF unlocks the standard DiT-B architecture (131M parameters), achieving a state-of-the-art FID of 3.37 where baselines fail to converge. We further showed that this geometric alignment scales efficiently to DiT-XL (FID 3.62) in just 80 epochs, establishing that efficient generation requires respecting the latent topology rather than just scaling model width.

\clearpage  


%
%
\bibliographystyle{splncs04}
\bibliography{main}


\section{Theoretical Derivation of Jacobi Field Regularization}
\label{theortical_derivation}
In this section, we provide the rigorous geometric derivation for the Jacobi Field Regularization shown in the main paper. We demonstrate that maximizing the fidelity of the generative trajectory requires weighting the learning objective by the metric distortion induced by the manifold's curvature.

\subsection{Geometric Setup}

Let $\mathcal{M} = \mathbb{S}_R^{d-1}$ be a hypersphere of radius $R$ embedded in $\mathbb{R}^d$. We define the flow trajectory $x_t$ traversing from data $x$ (at $t=0$) to noise $\epsilon$ (at $t=1$) along a geodesic path. The standard Riemannian Flow Matching (RFM) loss is defined in the tangent space $\mathcal{T}_{x_t}\mathcal{M}$:

\begin{equation}
    \mathcal{L}_{\text{RFM}} = \mathbb{E}_{t, x, \epsilon} \left[ \| v_{\theta}(x_t, t) - u_t(x_t|x, \epsilon) \|^2 \right]
\end{equation}

This objective treats velocity errors uniformly. However, due to the positive curvature of $\mathcal{S}^{d-1}$, the mapping from the tangent space at time $t$ to the endpoint $\epsilon$ is non-isometric. To ensure the generative flow precisely reaches the target noise manifold, we must minimize the error in the endpoint reconstruction rather than the instantaneous velocity.

The relationship between the tangent velocity $v$ at $x_t$ and the target endpoint $\epsilon$ is given by the Riemannian Exponential Map:
\begin{equation}
    \epsilon = \exp_{x_t}\left((1-t) \cdot \Omega \cdot \frac{v}{\|v\|}\right)
\end{equation}
where $\Omega$ is the total geodesic distance between $x$ and $\epsilon$. We define the Jacobi-Regularized loss as the squared distance in the target manifold space, weighted by the differential of this map.

\subsection{Jacobi Fields and Error Propagation}

To analyze the perturbation of the trajectory, we adopt the Jacobi field formulation from \cite{zaghen2025riemannian}. We consider a smooth family of geodesics $\{\gamma_s\}$ all starting from the same point $\gamma_s(0) := x_t \in \mathcal{M}$, determined by a perturbed initial velocity.

\textbf{Definition 4.1 (Jacobi field at a vanishing starting point).} Let the family of geodesics be defined as:
\begin{equation}
    \alpha(s, \tau) := \gamma_s : \tau \to \exp_{x_t} (\tau (v + s w))
\end{equation}
with $s \in [0,1]$ indexing the perturbation and $\tau \in [0,1]$ representing the affine parameter along the geodesic connecting $x_t$ to the endpoint $\epsilon$. Here, $v \in \mathcal{T}_{x_t}\mathcal{M}$ is the target velocity vector, and $w \in \mathcal{T}_{x_t}\mathcal{M}$ represents the error (perturbation) in the predicted velocity. 

For each fixed $\tau \in [0,1]$, the variation in the trajectory is described by the \textit{Jacobi field}:
\begin{equation}
    J(\tau) := \partial_s \alpha(s, \tau) \big|_{s=0}
\end{equation}
along the geodesic $\gamma_0(\tau)$. This field satisfies the Jacobi ODE:
\begin{equation}
    D_{\tau}^2 J + R(J, \dot{\gamma}_0)\dot{\gamma}_0 = 0
\end{equation}
where $R$ is the Riemannian curvature tensor. Following the initial conditions of the exponential map perturbation, the Jacobi field is uniquely defined by $J(0) = 0$ (vanishing error at the source $x_t$) and $D_{\tau}J(0) = w$ (the initial velocity error).

For a hypersphere $\mathbb{S}_R^{d-1}$ with constant sectional curvature $K = 1/R^2$, and denoting the total length of the geodesic segment from $x_t$ to $\epsilon$ as $L = (1-t)\Omega R$, the magnitude of the Jacobi field at any $\tau$ is given by the solution:
\begin{equation}
    \|J(\tau)\| = \|w\| L \frac{\sin(\sqrt{K} L \tau)}{\sqrt{K} L}
\end{equation}
Evaluating this at the endpoint $\tau=1$, which corresponds to the target noise $\epsilon$, yields the displacement error on the manifold.

\subsection{Derivation of the Weighting Factor $\lambda(t, \Omega)$}

We seek a weighting factor that scales the tangent space error $\|w\|^2$ to match the effective error at the target $\epsilon$. The Jacobian scaling factor $\mathcal{J}(t)$ is the ratio of the displacement at the target ($\tau=1$) to the linearized displacement (equivalent to Euclidean transport):

\begin{equation}
    \mathcal{J}(t) = \frac{\|J(1)\|}{\|w\| \cdot L} = \frac{\frac{1}{\sqrt{K}}\sin(\sqrt{K} L)}{L}
\end{equation}

On the hypersphere, substituting $\sqrt{K} = 1/R$ and noting that the arc length $L = R \cdot (1-t)\Omega$, the term $\sqrt{K}L$ simplifies to the remaining angular distance $(1-t)\Omega$. The expression thus becomes the normalized sinc function:

\begin{equation}
    \mathcal{J}(t) = \frac{R \sin((1-t)\Omega)}{R (1-t)\Omega} = \text{sinc}((1-t)\Omega)
\end{equation}

The regularized loss function minimizes the squared endpoint error, which corresponds to the squared norm of the Jacobi field at $\tau=1$. This requires the weight $\lambda(t, \Omega) = \mathcal{J}(t)^2$:

\begin{equation}
    \lambda(t, \Omega) = \text{sinc}^2((1-t)\Omega)
\end{equation}


\begin{algorithm}[t]
\caption{Train for RJF}
\label{alg:rfm_jacobi}
\begin{algorithmic}[1]
\REQUIRE Dataset $\mathcal{D}$, RAE feature Manifold $\mathcal{M} = \mathbb{S}^{d-1}$, Flow Model $v_\theta$, learning rate $\eta$, Logit-Normal parameters $\mu, \sigma$, Shift factor $s$
\WHILE{not converged}
    \STATE \quad Sample batch $x \sim \mathcal{D}$ and  $x \gets x / \|x\|$
    \STATE \quad Sample prior $\epsilon \sim \mathcal{N}(0, I)$ and  $\epsilon \gets \epsilon / \|\epsilon\|$
    \STATE \textbf{Time Sampling (Logit-Normal + Shift):}
    \STATE \quad Sample $t_{\text{raw}} \sim \text{LogitNormal}(\mu, \sigma)$ on $[0, 1]$
    \STATE \quad Apply Time Shift: $t \gets \frac{s \cdot t_{\text{raw}}}{1 + (s - 1)t_{\text{raw}}}$
    \STATE \textbf{Interpolate (SLERP):}
    \STATE \quad Compute geodesic distance $\Omega = \arccos(\langle \epsilon, x \rangle)$
    \STATE \quad $x_t = \frac{\sin((1-t)\Omega)}{\sin(\Omega)}x + \frac{\sin(t\Omega)}{\sin(\Omega)}\epsilon$
    \STATE \textbf{Target Velocity:}
    \STATE \quad $u_t = \dot{x}_t$ (projected to tangent space $T_{x_t}\mathcal{M}$)
    \STATE \textbf{Jacobi Weighting:}
    \STATE \quad $w_t = \left( \frac{\sin((1-t)\Omega)}{(1-t)\Omega} \right)^2$
    \STATE \textbf{Loss Computation:}
    \STATE \quad $\hat{v} = v_\theta(x_t, t)$
    \STATE \quad $\hat{v}_{\text{proj}} = \hat{v} - \langle \hat{v}, x_t \rangle x_t$
    \STATE \quad $\mathcal{L} = w_t \cdot \| \hat{v}_{\text{proj}} - u_t \|^2$
    \STATE Update $\theta \gets \theta - \eta \nabla_\theta \mathcal{L}$
\ENDWHILE
\end{algorithmic}
\end{algorithm}

\begin{algorithm}[t]
\caption{Sampling for RJF}
\label{alg:rfm_sampling_geo}
\begin{algorithmic}[1]
\REQUIRE Trained Flow Model $v_\theta$, Steps $N$, Class Label $y$, Latent Dimension $d$, Target Radius $R$, Shift factor $s$
\STATE \textbf{Initialization:}
\STATE \quad Sample prior $\epsilon \sim \mathcal{N}(0, I)$
\STATE \quad Project to sphere: $x \gets \epsilon / \|\epsilon\|$
\STATE \quad Sample $t_{\text{raw}} \sim \text{LogitNormal}(\mu, \sigma)$ on $[0, 1]$
\STATE \quad Apply Time Shift: $t \gets \frac{s \cdot t_{\text{raw}}}{1 + (s - 1)t_{\text{raw}}}$
\FOR{$i = 0$ to $N-1$}
    \STATE \quad Current time $t \gets t_i$, \quad Next time $t' \gets t_{i+1}$
    \STATE \quad Step size $\Delta t \gets t' - t$
    \STATE \quad Predict velocity: $v \gets v_\theta(x_{\text{in}}, t, y)$
    \STATE \quad Remove radial component: $v_{\text{tan}} \gets v - \langle v, x \rangle x$
    \STATE \quad Calculate angle: $\theta \gets \|v_{\text{tan}}\| \cdot \Delta t$
    \STATE \quad Update position via rotation:
    \STATE \quad $x \gets \cos(\theta) x + \sin(\theta) \frac{v_{\text{tan}}}{\|v_{\text{tan}}\|}$
    \STATE \quad Re-normalize: $x \gets x / \|x\|$
\ENDFOR
\STATE \textbf{Final Output Scaling:}
\STATE \quad $x_{\text{out}} \gets x \cdot R$
\RETURN $x_{\text{out}}$
\end{algorithmic}
\end{algorithm}

\begin{figure*}[t]
    \centering
    \includegraphics[width=1.0\linewidth]{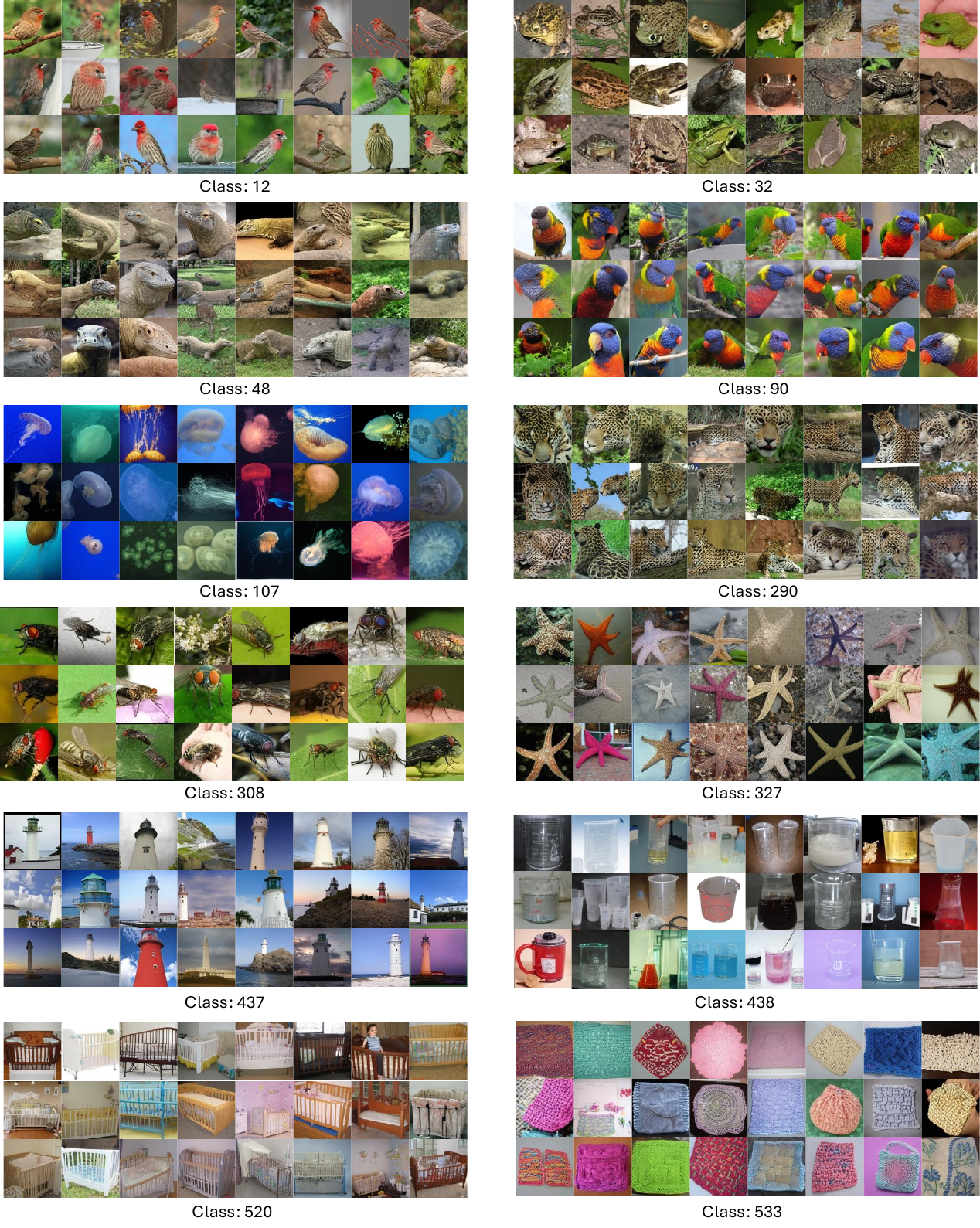}
    \vspace{-20pt}
    \caption{\textbf{Qualitative results of LightingDiT-XL+RJF trained for 80 epochs on ImageNet 256$\times$256.} We show uncurated results on the 12 classes .}
    \label{fig:supplementary_1}
\end{figure*}

\begin{figure*}[t]
    \centering
    \includegraphics[width=1.0\linewidth]{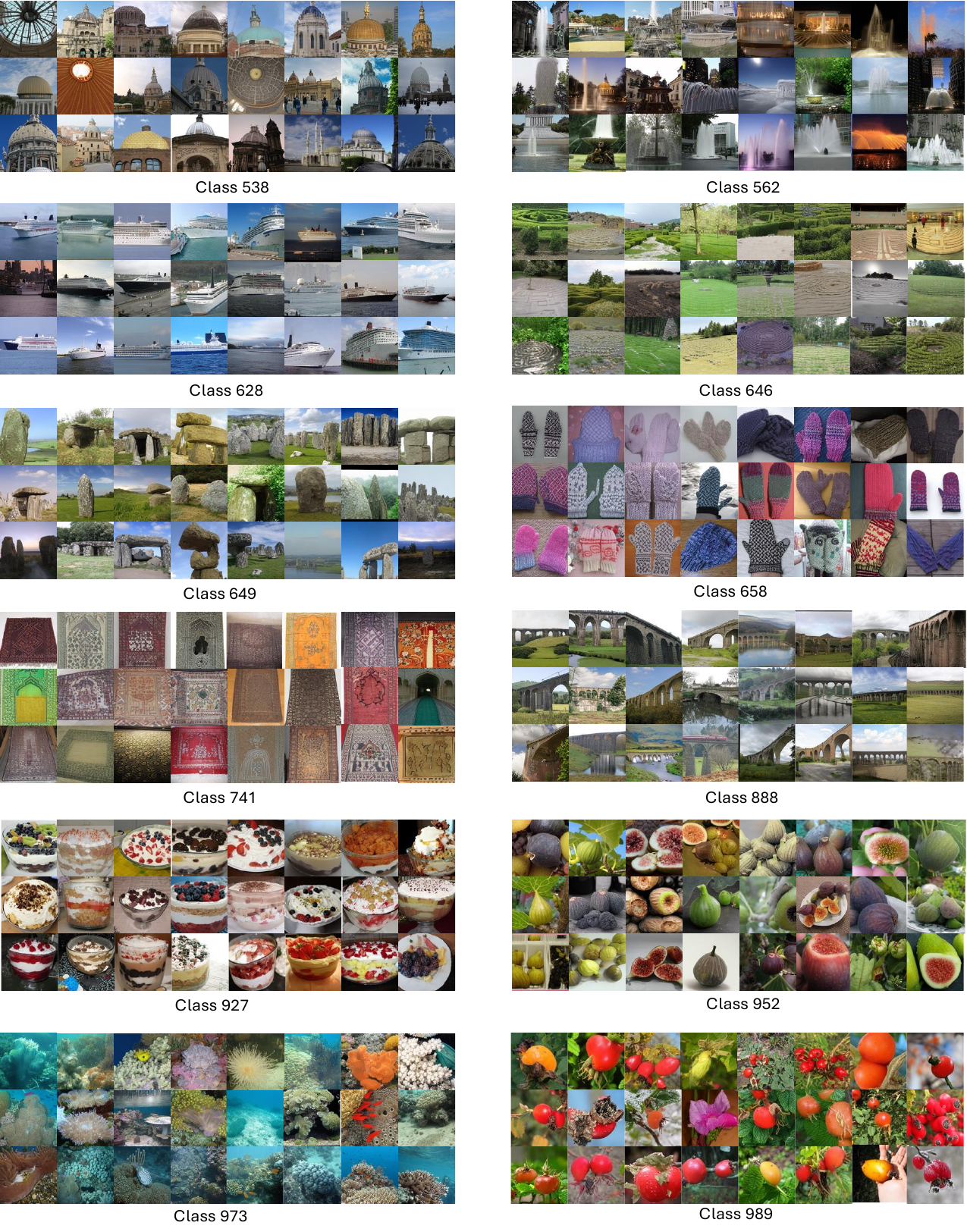}
    \vspace{-20pt}
    \caption{\textbf{Qualitative results of LightingDiT-XL+RJF trained for 80 epochs on ImageNet 256$\times$256.} We show uncurated results on the 12 classes .}
    \label{fig:fig:supplementary_2}
\end{figure*}

\clearpage

\end{document}